\documentclass[journal]{IEEEtran}

\usepackage{times}
\usepackage{epsfig}
\usepackage{graphicx}
\usepackage{amsmath}
\usepackage{amssymb}
\usepackage{subfigure}
\usepackage{multirow}
\usepackage{array}
\usepackage{color}
\usepackage{booktabs}
\usepackage{caption}
\ifCLASSINFOpdf
\else
\fi
\begin{document}
%
\title{LGA-RCNN: Loss-Guided Attention for Object Detection}
%
%
%

\author{Xin~Yi$^*$,
	Jiahao~Wu$^*$,
	Bo~Ma$^\dagger$,
	Yangtong~Ou,
	Longyao~Liu
	\thanks{$^*$Equal contribution.}
	\thanks{$^\dagger$Corresponding author.}}

\maketitle

\begin{abstract}

Object detection is widely studied in computer vision filed. In recent years, certain representative deep learning based detection methods along with solid benchmarks are proposed, which boosts the development of related researchs. However, existing detection methods still suffer from undesirable performance under challenges such as camouflage, blur, inter-class similarity, intra-class variance and complex environment. To address this issue, we propose LGA-RCNN which utilizes a loss-guided attention (LGA) module to highlight representative region of objects. Then, those highlighted local information are fused with global information for precise classification and localization.
\end{abstract}


%
\IEEEpeerreviewmaketitle

\section{Introduction}
%
%
%
%
%
%

\IEEEPARstart{O}{bject} detection is a fundamental problem in computer vision, which can be applied in instance segmentation, scene understanding, pose estimation, image captioning and multiple objects tracking (MOT), to name a few. Given an arbitrary image, the goal of object detection is to determine the presence of the predefined categories and locate them in this image. Recently, with the development of convolutional neural network, learning based object detection methods have achieved remarkable progress beyond the traditional detection methods. Meanwhile, in order to train and evaluate the performance of different detection models, certain solid benchmarks for object detection have also been proposed by researchers.

The state of the object is actually complicated (camouflage, occlusion or high-speed state), which brings challenges to object detection methods. Those challenges include (1) Complex environment: objects is obscured by smoke or flames; (2) Intra-class variance: the appearance of the same category could be quite different; (3) Inter-class similarity: the appearance of the different categories could be quite similar; (4) Scale: objects at different distances would generate scale differences; (5) Motion blur: objects are usually in motion; (6) Camouflage: objects are decorated with camouflage. Therefore, existing object detection methods suffer from undesirable performance.

In this work, we propose a Loss-Guided Attention RCNN (LGA-RCNN) to tackle those challenges by highlighting representative region. We find that in dense detection framework, RoI module can generate almost all features of foreground objects and the bottleneck of performance lies in the classification of RoI features. Thus, we append a LGA module behind RoI feature layers, which predicts $k$ Gaussian masks on RoI feature maps to seek discriminative parts of objects for more accurate classification. In addition, an extra classification loss is imposed on masked RoI feature maps to ensure that those Gaussian masks converge to optimal locations. Compared with common attention modules like CBAM~\cite{woo2018cbam} which only focus on contextual information (rather than global information), our method makes full use of global information to mine representative local parts. Besides, time and memory consumption of our method are also better than global-range methods like non-local~\cite{wang2018non}.

Our contributions can be summaried as follows.

We propose LGA-RCNN which utilizes a loss-guided attention (LGA) module to highlight representative region of objects and improve detection performance.

\section{Related Works}

\subsection{Datasets}

Datasets play a very important role in the history of learning-based object detection methods. Previous detection datasets can be divided into single-category object datasets and multi-category object datasets (general object datasets). Single-category object dataset only contains one specific category of object such as face~\cite{fddbTech, faceevaluation15,klare2015pushing,yang2016wider}, pedestrian~\cite{dollar2011pedestrian,zhang2017citypersons,zhang2019widerperson}, vehicle~\cite{cordts2016cityscapes}, apple~\cite{hani2020minneapple}, etc. Multi-category object dataset contains multiple types of objects such as person, bicycle or car. Previous  representative works of multi-category object datasets include ImageNet~\cite{deng2009imagenet}, PASCAL VOC 2007~\cite{everingham2010pascal}, PASCAL VOC 2012~\cite{everingham2015pascal}, MS COCO~\cite{lin2014microsoft} and Open Images~\cite{kuznetsova2018open}. Specifically, the detailed information of each dataset is listed in Table~\ref{comparsion of dataset}.

\begin{table}[t]
	\caption{Comparison of Object Detection Benchmarks}
	\resizebox{0.49\textwidth}{!}{
		\begin{tabular}{ccccccc}
			\toprule
			Dataset    & Specific Field & Categories & Boxes/Images \\ \midrule
			Pascal VOC~\cite{everingham2015pascal} & Not Specific   & 20         & 2.4       \\
			ImageNet~\cite{deng2009imagenet}   & Not Specific   & 200        & 1.1       \\
			COCO~\cite{lin2014microsoft}       & Not Specific   & 80         & 7.3       \\
			OpenImages~\cite{kuznetsova2018open} & Not Specific   & 600        & 9.3       \\ \bottomrule
		\end{tabular}%
	}
	\label{comparsion of dataset}
\end{table}

Although those datasets show their effectiveness under the verification of numerous algorithms, they are collected for generic object detection, in which the types of objects are broad but not specialized. The dataset for a specific field is necessary because the characteristics of objects in different fields are quite different. And detection methods in specific field need to be improved to adapt to these characteristics, such as apple detection using enhanced YOLO-v3~\cite{tian2019apple}. Thus, a robust detection algorithm is quite necessary.

\subsection{Methods}

According to whether to utilize region proposal, object detecion methods can be divided into two mainstreams, two-stage methods and one-stage methods.

\subsubsection{Two-Stage Methods}

Similar to tranditional object detection methods, two-stage object detection methods utilize a region porposal stage to generate sufficient candidate regions. 


Inspired by selective search~\cite{uijlings2013selective}, Girshick~\cite{girshick2014rich} proposes RCNN in 2014 for generic object detection. However, repetitive feature extraction in RCNN causes slow operation. Thus, He et al.~\cite{he2015spatial} propose SPPNet to reduce calculation time by obtaining proposals from the whole feature maps rather than the whole source image. Besides, Fast RCNN~\cite{girshick2015fast} is proposed with a Region of Interest (RoI) pooling layer to generate proposals of the same scale. Networks behind RoI layer become end-to-end so that detection speed is accelerated. Moreover, Ren et al.~\cite{ren2015faster} replace selective search with Region Proposal Network (RPN) in Faster RCNN, which sets $k$ anchors with different aspect raito in feature maps to generate proposals.

Recently, more two-stage methods~\cite{dai2016r, li2017light, gkioxari2019mesh, lu2019grid, beery2020context} are  proposed to enhance speed and performance. However, due to the existence of RoI, the speed of the two-stage method is still slow and cannot meet the requirements of real-time detection. Thus, one-stage methods are proposed.

\subsubsection{One-Stage Methods} 

Unlike two-stage methods, one-stage methods achieve object detection without a distinct region proposal stage. According to whether to utilize anchor, they can be further devided into anchor-based methods and anchor-free methods.

Anchor-based one-stage methods apply anchors to classify object category directly rather than to generate region proposals. Liu et al.~\cite{liu2016ssd} propose a fully convolutional network SSD, which sets anchors in features with multiple scale to achieve detection on objects with different size. Then, Kong et al.~\cite{kong2017ron} propose enchanced SSD algorithm, RON, that adds multiple deconvolutional layers to improve the detection capability in small objects. Lin et al.~\cite{lin2017focal} propose RetinaNet with 9 anchors in each FPN scale. This work also introduces the focal loss to solve the imbalance between positive sample assignment and negative sample assignment.

Those anchor-based one-stage methods are dependent on the setting of the anchor parameters to a large extent and unreasonable configuration prevents the anchor box from matching the target box well, resulting in performance drop. Thus, anchor-free one-stage methods are proposed~\cite{redmon2016you, law2018cornernet, zhou2019objects, zhou2019bottom}. Specifically, YOLO~\cite{redmon2016you} regards the object detection problem as the regression problem, where the feature map is split into $S\times S$ grid cells and each cell is responsible for preditcing objects centered at this cell. CornerNet~\cite{law2018cornernet} and CenterNet~\cite{zhou2019objects} convert object detection problem into a keypoint detection problem. Besides, ExtremeNet~\cite{zhou2019bottom} utilizes the labeled data in the segmentation dataset to predict the boundary points and the center point of the object. The boundary points are guaranteed to fall into foreground area, so they are easier to detect than corner points. However, this algorithm needs to be trained with the mask annotation, increasing the acquisition cost.

\section{LGA R-CNN}
\label{method}

As illustrated above, several challenges exist in object detection, e.g., occlusion, camouflage, and complex environment, which causes the performance drop to some degree. Thus, targeting at addressing this issue, we propose LGA R-CNN for object detection. 

\begin{figure*}[t]
	\begin{center}
		\includegraphics[width=0.9\textwidth]{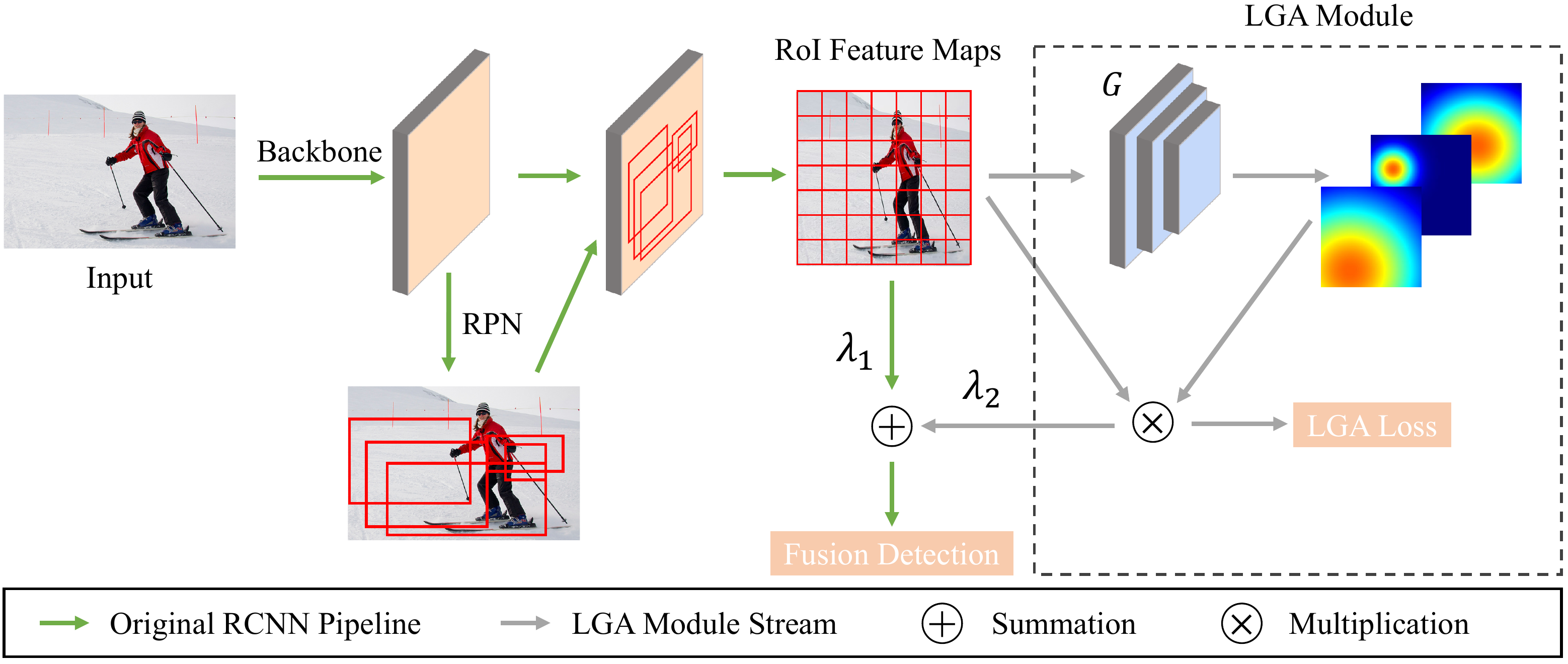}  
	\end{center}
	\caption{Illustration of our proposed LGA R-CNN. Foreground object feature is extracted by Region Proposal Network (RPN). Then, we utilize a Loss-Guided Attention (LGA) module to predict several Gaussian maps from object feature and highlight discriminative part of feature map using those predicted Gaussian maps. LGA module is supervised and guided by classification loss under highlighted feature map. Furthermore, in order to achieve better regression performance, we fuse global information (original feature) and local information (highlighted feature) for final classification and regression.}
	\label{fig:framework}
\end{figure*}

\subsection{Overall}

We build our method LGA-RCNN based on R-CNN framework and the whole pipeline is illustrated in Figure~\ref{fig:framework}. Given an arbitrary image, RCNN detector utilizes backbone network and region proposal network (RPN) to generate feature maps with certain proposals. Then, RoI align is applied to crop RoI feature maps from the whole image feature maps. In such a dense detection framework, the bottleneck of performance lies on networks behind RoI features. Thus, besides the common classification and regression branches, we append auxiliary LGA module on RoI feature maps to predict and highlight representative regions for more accurate classification. Afterwards, those highlighted features are fused with the original RoI feature for preciser classification and regression.

\subsection{LGA Module}

The principal of designing LGA module is to mine and highlight those more representative and discriminative regions of the object, and reduce the adverse effect in potential region with occlusion, camouflage or other interference. To achieve this target, the proposed component should be able to sense the global information and seek the local region with more discriminative clues. Thus, we utilize a network to predict the Gaussian attention masks from the global RoI features. Assuming that those representative regions should be discriminative enough for a detector to do the classification, e.g., a person's face is strong enough to be distinguished from other categories, we attach a classification loss to force LGA to learn a better attention. Furthermore, original global information need to be maintained for accurate locating and classification fine-tuning. Thus, we fuse those masked local-enhanced feature maps with those original feature maps for final detection heads.


\subsubsection{Gaussian mask prediction}

Common attention module such as CBAM~\cite{woo2018cbam} is implemented with channel-wise pooling and spatial-wise convolution, which thus leads to the lack of the global information. Non-Local methods are able to percept global information, but they are much more complicated and time-consuming. In LGA module, we construct a learnable mapping function to map the global features into a Gaussian parameters ($\mu$ and $\sigma$) then transfer those parameters into Gaussian masks. To be specific, given RoI feature $x$ with $256$ channels and $7\times7$ spational resolution, we first downsample the feature into a lower channel dimension to avoid high complexity by network $f_{d}$. Then, network $f_{c}$ is applied on the downsampled feature to predict Gaussian parameters.
\begin{equation}
	\begin{aligned}
		\mu &= S_{ratio}*Tanh(f_{c}(f_{d}(x))) \\
		\sigma &= ReLU(f_{c}(f_{d}(x))) + 1
	\end{aligned}
\end{equation}
We utilize $Tanh$ and $S_{ratio}$ to ensure that $\mu$ falls in the range of the spatial resolution of feature ($[0,7]$ in this case) and $ReLU$ to ensure $\sigma$ is no less than $1$. Actually, Gaussian parameters are capable of representing some instance-level semantic information. For a RoI region of $7\times7$ size, the way we obtain gaussian parameters ensure that it can sense high-level semantic feature of the target instance. $\mu$ can be regarded as a position prediction on the discriminative region, while $\sigma$ can be regarded as the scale of this region.

\subsubsection{Loss-Guided Training}

After initialized, different Gaussian masks pay attention to different regions, i.e., different local features are enhanced. We hope that those Gaussian masks would focus on more representative and discriminative regions. For example, when it comes to a picture with excavators and vehicles, those unique parts such as caterpillar tread is more discriminative than similar parts like steel shell. To achieve this, we apply an extra classification loss on masked RoI feature maps for supervision. Assuming that common attention module do benefit the performance where they probably focuses on the steel shell, however the highlighted feature could be a disadvantage to distinguish excavators out of vehicles. Loss-Guided training attention is designed to focus on a more discriminative region like barrel, which would not be a part of the vehicles. With the supervision of the classification loss on Gaussian feature, the LGA module is forced to search for the aforementioned region to make the new-attached loss decline. 


\subsubsection{Feature Fusion}

Although classification accuracy is improved by enhanced local informaion, part of global information is sacrificed in those highlighted RoI features. Therefore, inaccurate position regression would appear if we directly using highlighted features to locate the object. In order to maintain the accuracy and robustness of the bboxes regression process, we fuse masked RoI feature maps with original RoI feature maps to combine local information with global information. Then, we apply final detection on fused RoI features. Furthermore, part of Gaussian mask focuses on marginal region of the object. Thus, fused RoI features can sense more on the outline of the objects, which enhances the result of location.

\section{Conclusion}

In this work, we analyze certain challenges in object detection including camouflage, motion blur, compliated environment, intra-class variance, inter-class similarity and scale. Then, we propose the Loss-Guided Attention RCNN (LGA-RCNN) to address those issues by adding LGA module in common R-CNN framework. LGA module utilizes a network to predict Gaussian masks from RoI features and force those masks to focus on representative regions of object by an extra LGA loss.

\newpage


%





\ifCLASSOPTIONcaptionsoff
  \newpage
\fi



%
%
%

\bibliographystyle{./IEEEtran}
\bibliography{./IEEEabrv,./IEEEexample}

%








\end{document}